%% file: eccv22_sectioned.tex
\documentclass[runningheads]{llncs}
\usepackage{graphicx}
\usepackage{algorithm}
\usepackage{algpseudocode}

\usepackage{tikz}
\usepackage{comment}
\usepackage{amsmath,amssymb} 
\usepackage{color}
\usepackage{subcaption}
\usepackage{booktabs} 
\usepackage{multirow}
\usepackage{tablefootnote}
\usepackage{enumitem}
\usepackage{subcaption}
\usepackage[accsupp]{axessibility}  
\usepackage[pagebackref=true,breaklinks=true,colorlinks,bookmarks=true]{hyperref}
\usepackage[numbers,sort&compress]{natbib}
\usepackage{float}
\floatstyle{plaintop}
\restylefloat{table}
\usepackage{amsmath}

\definecolor{mydarkblue}{rgb}{0,0.08,0.45}
\definecolor{urlcolor}{rgb}{0,.145,.698}
\definecolor{linkcolor}{rgb}{.71,0.21,0.01}
\hypersetup{ %
	pdftitle={},
	pdfauthor={},
	pdfsubject={},  
	pdfkeywords={},
	pdfborder=0 0 0,     
    pdfpagemode=UseOutlines,
	breaklinks=true,  
	colorlinks=true,
	bookmarksnumbered=true,
	urlcolor=urlcolor,
	linkcolor=linkcolor,
	citecolor=mydarkblue,
	filecolor=mydarkblue,
	pdfview=FitH}
	
\vbadness=\maxdimen
\newcommand{\methodname}{BD-BNN}

\begin{document}
\pagestyle{headings}
\mainmatter
\def\ECCVSubNumber{5701}  

\title{Bi-Modal Distributed Binarized Neural Networks } 

\titlerunning{Abbreviated paper title}
%
\author{Tal Rozen\inst{1}  \and
Moshe Kimhi\inst{1} \and
Brian Chmiel\inst{1,2} \and
Avi Mendelson\inst{1} \and
Chaim Baskin\inst{1}}
\authorrunning{F. Author et al.}
%
\institute{Technion -- Israel Institute of Technology\\\email{\{tal.rozen,moshekimhi,brianch,mendlson,chaimbaskin\}@technion.ac.il} \and
Habana Labs  --  An Intel company, Caesarea, Israel
\\
}
\maketitle

Binary Neural Networks (BNNs) are an extremely promising method to reduce deep neural networks' complexity and power consumption massively. Binarization techniques, however,  suffer from ineligible performance degradation compared to their full-precision counterparts.

Prior work mainly focused on strategies for sign function approximation during forward and backward phases to reduce the quantization error during the binarization process. In this work, we propose a Bi-Modal Distributed binarization method (\methodname{}). That imposes bi-modal distribution of the network weights by kurtosis regularization. The proposed method consists of a training scheme that we call Weight Distribution Mimicking (WDM), which efficiently imitates the full-precision network weight distribution to their binary counterpart. 
Preserving this distribution during binarization-aware training creates robust and informative binary feature maps and significantly reduces the generalization error of the BNN. Extensive evaluations on CIFAR-10 and ImageNet demonstrate the superiority of our method over current state-of-the-art schemes. Our source code, experimental settings, training logs, and binary models are available at \url{https://github.com/BlueAnon/BD-BNN}.

\input{sections/Introduction}
\input{sections/Related_Work}
\input{sections/Preliminaries}

\input{sections/Method}
\input{sections/Experiments}
\input{sections/Conclusions}

\clearpage
%
%
\bibliographystyle{splncs04}
\bibliography{egbib}

\clearpage

\appendix{}

\renewcommand\thefigure{\thesection.\arabic{figure}} 
\renewcommand\thetable{\thesection.\arabic{table}} 
\renewcommand\theequation{\thesection.\arabic{equation}}  
\setcounter{figure}{0}  
\setcounter{table}{0}


\input{sections/appendix.tex}

\end{document}

%% file: sections/Introduction.tex
\section{Introduction}

\begin{figure}[!h]
	\centering
	\includegraphics[width=1.0\textwidth]{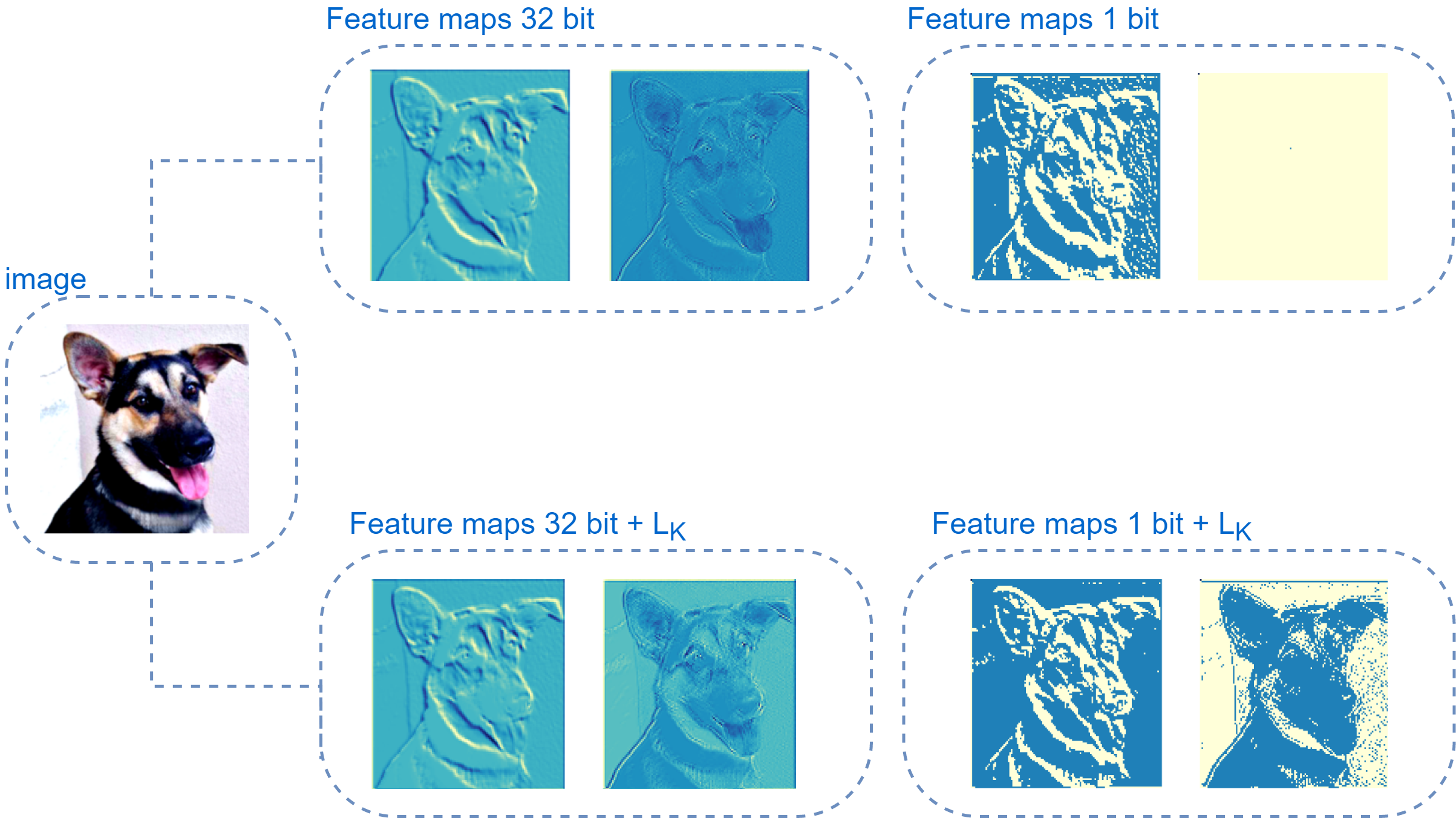} 
	\vspace{-0.3cm}
	\caption{Feature maps taken from the first layer of ResNet-18 trained on an ImageNet dataset. Top row: 32-bit features and binary features from a pre-trained neural network, bottom row: features trained on ImageNet with the proposed weight distribution regularization ($L_K$). 
	While 32-bit feature maps are quite similar, the binary representation of these features lack expressiveness and are less informative.}
    \label{fig:activation_visualization}
\end{figure}

Deep neural networks (DNNs) are widely used nowadays for a variety of tasks, including computer vision \cite{resnet,Liu2022ACF,Zou2019ObjectDI}, natural language processing \cite{Vaswani2017AttentionIA}, control of autonomous vehicles \cite{Huang2020AutonomousDW} and more . The large size of most such models, however, makes them expensive (in terms of hardware and inference time) to deploy on devices with small computational and memory resources. To cope with the huge demand in resources made by DNNs, academic and industrial researchers have attempted to compress these models and relax the inference cost. The leading compression methods includes pruning \cite{Frankle2019TheLT,Hubara2021AcceleratedSN}, low-rank decomposition \cite{Chmiel2020FeatureMT,Baskin2021CATCT}, quantization \cite{banner2018post,Chmiel2020NeuralGA} and knowledge distillation \cite{Hinton2015DistillingTK}.

In quantization, the goal is to reduce the numerical precision of the DNNs parameters and activations, which in the full-precision version are commonly represented by 32 bits. In the extreme case, they can be reduced to 1-bit representation, which means getting a 32x memory footprint reduction and a higher computational reduction by transforming the expensive matrix computation into cheaper XNOR and popcount operations \cite{Rastegari2016XNORNetIC}.

Notwithstanding the immense interest in BNNs, the challenge today to close the accuracy gap between the full precision (FP) and  binarized versions yet to be completed. The importance of maintaining the appropriate data distribution is a key issue in many works, specifically in DNN compression studies. While \cite{banner2018post} uses a normal distributed prior for the weights and activations, \cite{Chmiel2020NeuralGA} uses a log-normal distribution prior for the neural gradients. In both cases, they show that choosing a different distribution approximation, induces a significant degradation in accuracy. KURE \cite{chmiel2020robust} represents a step forward by offering matching between the data distribution and the task. The authors show that uniform distribution is more robust under quantization noise and design a method to uniformize the distribution of the weights. 

In this paper, we suggest a method to manipulate the weight distribution in order to adopt a bi-modal distribution. The manipulation is done with a unique combination of a bi-modal regularization term, and a novel training regime of weight regularization mimicking alongside knowledge distillation (KD). This distribution is capable of reducing the quantization error under model binarization. Our experiments show state-of-the-art results in a variety of BNN without increasing the number of parameters. For example, we achieved a 60.6\% accuracy in the standard binarized ResNet-18 on ImageNet dataset compared with 60.2\%, the previous state-of-the-art result.

This paper makes the following contributions: 
\begin{itemize}
    \item We introduce a regularization term to manipulate the weight distribution such that it becomes a bi-modal distribution that, in turn, reduces the quantization error.
    \item We propose WDM as a training scheme that establishes a better match between the teacher and student distribution for an FP 32 teacher and binary student.
    \item We are the first to analyze the distribution of the gradients of the loss term with regularization for BNNs.  
    \item We analyze the proposed method in a variety of BNNs, and achieve state-of-the-art results without increasing the number of parameters.
\end{itemize}

%% file: sections/Related_Work.tex
\section{Related Work}

Starting from the pioneering binarization work \cite{Courbariaux2016BinarizedNN}, which showed decent performance for 1-bit quantization of the forward pass adopting the straight-through-estimator (STE)\cite{Bengio2013EstimatingOP} to approximate the binarized gradients, a lot of research has been applied to reduce the gap between binarized DNNs and their FP counterparts. To reduce the quantization error, XNOR-Net\cite{Rastegari2016XNORNetIC} uses two different scaling factors for the weights and activations. Bi-Real-Net \cite{Liu2018BiRealNet} combines additional shortcuts and a replacement of the sign function with a polynomial function for the activations to reduce the information loss caused by the binarization.  IR-Net \cite{Qin2019IRNetFA} combines a technique to reduce the information entropy loss in the forward pass with an error decay estimator (EDE) that replaces the STE in the backward pass. 

Real-to-Bin \cite{Martnez2020TrainingBN} suggests changing the order of the ResNet block, using a two-stage optimization strategy and introduces a progressive teacher-student technique. SD-BNN \cite{Xue2021SelfDistributionBN} proposes removing the commonly used scaling factor for weights and activations, and replaces them with a self-distribution factor that is learnable and applied before the binarization. RBNN \cite{Lin2020RotatedBN} explores the effect of the angular bias on the quantization error and suggests a method that reduces the angle between the FP weights and their binarized version. They also suggests a new back-propagation method that meets their needs more efficiently. \cite{Bulat2019ImpTrain} combine a reverse-order initialization with smooth progressive quantization and binarized KD. They use the output heatmaps of the teacher network as soft labels for the binary cross entropy loss. FDA \cite{Xu2021LearningFD} suggests estimating the gradient of the sign function in the Fourier frequency domain. 

A different approach taken by other related works is to consider the DNN's distribution to improve performance. KURE 
\cite{chmiel2020robust}
applies a regularization to the weights to make them uniformly distributed, which improves the network robustness. Both in \cite{banner2018post} and \cite{Chmiel2020NeuralGA} a prior distribution is used to reduce the quantization error. while the former use a normal distribution prior for the weights and activations and the latter use a log-normal distribution prior for the neural-gradients.

In this work, we focus on DNN binarization that does not change the number of parameters of the original network, in contrast to recent known methods, such as ReActNet \cite{Liu2020ReActNetTP}, that show impressive binarization results at the price of increasing the number of parameters.

%% file: sections/Preliminaries.tex
\section{Preliminaries}

In this section we review and provide the notation for BNN, kurtosis and KD.

\subsection{BNNs}
DNNs usually comprise two integral parts.
One is the feature extractor, and the other is a regressor or classifier. 
The feature extractor commonly uses a basic block based on some multiplication of weights and activations, such as a convolution layer, since the convolution operation is shift invariant and the structure of the network creates informative features for the learning task.
The convolution operation is defined as
$$Z_i = A_i \ast W_i$$ 
where $A_i \in \mathbb{R}^{K_i,L_i}$ is a two-dimensional input signal (also called activation) to the convolution and $W_i = {W_{i,j}...}_{c_i}$ is a set of $c$ two-dimensional weights $W_{i,j} \in \mathbb{R}^{M_i,N_i}$, both represented in floating-point over 32 bits. They also both use floating-point Multiply-ACcumulate (MAC) operations. A layer uses $c \times (K_i-M_i) \times {L_i-N_i}$ times $M_i \times N_i$ MAC operations.
One can see that CNNs have very low speed and high memory consumption due to the hardware computational and communication limitations.
BNNs, on the other hand, quantize the weights and activations to  1-bit representation and use 1-bit operations, simply by defining the binary projections using the sign function.
$$B_x=sign(x)=\begin{cases}
+1 , & x\geq0\\
-1, & else
\end{cases}$$
where $x$ is the floating-point parameter and $B$ is its binary equivalent. $B_w$ and $B_a$ denote the binary weight and activations, respectively. Accordingly, for BNNs, the convolution operation can be expressed as:
$$Z=I \ast W \approx B_w \ast B_a = (B_w \otimes B_a) \cdot \alpha$$
where $\otimes$ is the bitwise XNOR operation and $\alpha$ is a 32-bit scaling factor that is added to minimize the quantization error.
BNNs enable the user to reduce the latency and increase the throughput, as they now have a 32-times lower memory footprint and thus have a reduced access to memory needed. This is in addition to operations that are 58 times faster  compared with 32-bit MAC operations \cite{Rastegari2016XNORNetIC}.

\subsection{Knowledge distillation}
KD is a technique for transferring knowledge from one model, referred as a teacher, to another, referred as a student \cite{Hinton2015DistillingTK}.
Generally, a large DNN serves as a teacher and a more compact model serves as a student. 

The knowledge transferring is done by adding a distillation loss to the student target: 
\begin{equation}\label{KD_loss}
    \mathcal{L}_{KD} = D_{KL}(z_T || z_S)
\end{equation}
where $D_{KL}$ refers to Kullback–Leibler divergence and 
$z_T$ and $z_S$ are the output of the last fully connected layer in the teacher and student network, respectively.

KD is a very common technique for improving the generalization of the student \cite{AllenZhu2020TowardsUE}.
Some studies show improvements even when using the exact same model for the teacher and the student \cite{AllenZhu2020TowardsUE,Mobahi2020SelfDistillationAR,Zhang2019BeYO}. This regime is typically called self-distillation.
KD also leads to a tremendous improvement in model compression \cite{Hinton2015DistillingTK,Bucila2006ModelC,Zhang2019BeYO}, when the common setup for quantization is to use the FP model as teacher and the quantized model as a student.\cite{Kim2019QKDQK,Polino2018ModelCV}.

\subsection{Kurtosis}
Let X be a random variable, s.t. $X \in \mathcal{R}^d$ from a known probability distribution function (PDF). A common statistical analysis of a variable distribution is to look at the moments of the variable and the most common moments are the first and second moments $\mu[X] = \mathbb{E}(x)$) and ${\sigma}^2[X] = \mathbb{E}[(X-\mathbb{E}(x))^2]$.
A deeper analysis of the distribution can be done using the third and fourth moments -- skewness and kurtosis.

The kurtosis of a random variable is defined as follows:
\begin{equation}\label{eq:kur}
Kurtosis\left[X\right] = \left[ \mathbb{E} \left( \frac{X-\mu}{\sigma} \right) ^4\right]
\end{equation}
where $\mu$ and $\sigma$ are the first and second moments of X.\newline

The Kurtosis, also called 'tailedness', is a scale and shift-invariant measurement that helps explain how much of the PDF is dense around the mean.
For example, for a uniform distribution, kurtosis = 1.8 and for a distribution with that kurtosis, there is an equal possibility of being far and close to the mean.
In this paper, we use the kurtosis as a proxy of the weight distribution and define a kurtosis loss term that, by minimizing it, helps us mimic a desired distribution.
We show that subtracting the cumulative distribution function (CDF) of this distribution from the CDF of the network weights (i.e., distribution shifting) leads to a distribution with a bi-modal notion that is optimal for binarization.

%% file: sections/Method.tex
\section{Method}

In this section we first explore the effects of the fourth moment on the weight distribution. We demonstrate in Figure \ref{fig:activation_visualization} how the changes in distribution affect the network's feature maps. Then we discuss KD technique that supports our goal of achieving bi-model distributions that are more "binary friendly". Lastly, we present our Bi-Modal Distribution aware training method.  

\subsection{Weight distribution regularization}
\label{method:regul}
DNN's weights and activations usually follow a Gaussian or Laplace distribution  \cite{banner2018post}. For BNNs, this distribution may not be optimal. We would like to manipulate the distribution without harming the performance, so we can actually reduce the quantization error and thus increase the performance.  
In this work, we follow the work done by \cite{chmiel2020robust} and use the kurtosis as a proxy for the probability distribution. 
Kurtosis regularization is applied to the model loss function, $\mathcal{L}$, as follows:
\begin{equation}\label{eq:loss}
\mathcal{L} = \mathcal{L}_p + \lambda\mathcal{L}_K 
\end{equation}
where $\mathcal{L}_p$ is the target loss function. In common classification and regression tasks, $\mathcal{L}_p$ represents the cross entropy (CE) and mean squared error (MSE) loss, respectively.
$\mathcal{L}_K$ is the kurtosis term and $\lambda$ is the coefficient. $\mathcal{L}_K$  is defined as
\begin{equation}\label{eq:kur_loss}
\mathcal{L}_K = \frac{1}{L}\sum_{i=1}^{L}\left|Kurtosis\left[W_i\right]-K_T\right|^2
\end{equation}
where $L$ is the number of layers and $K_T$ is the target of the kurtosis regularization.

\begin{figure}[!h]
	\centering
	\includegraphics[width=1.\textwidth]{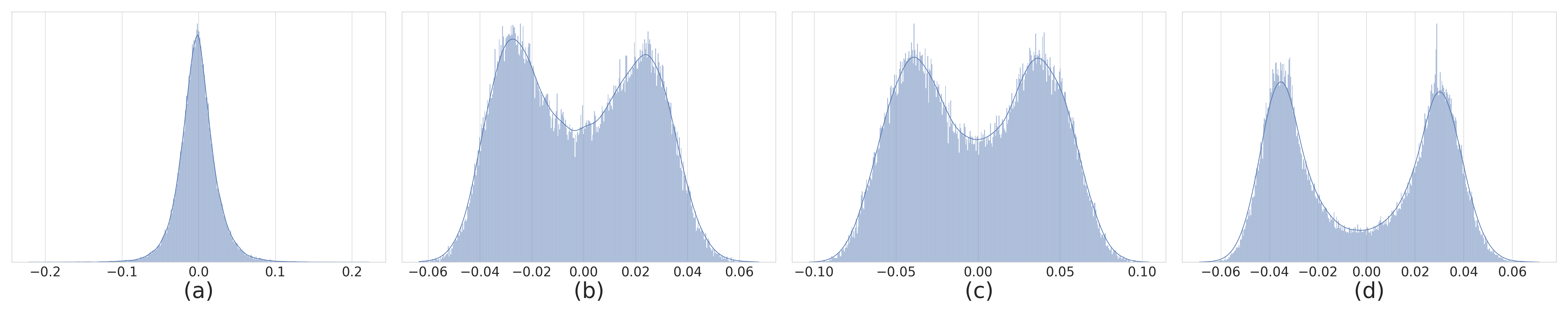} 
	\vspace{-0.3cm}
	\caption{An example of the distribution of a filter from the second layer of ResNet18 trained with the CIFAR10 dataset. We see the original weight distribution (a), the distribution with kurtosis loss with $K_T=1.8$ (b), with $K_T=1$ (c) and with $K_T$ that are set differently for each layer of the network (d)}
	\label{fig:hetro_kt}
\end{figure}

In this work we explore the effect of the gradients of $\mathcal{L}_K$ in general, and the value of $K_T$ on the model weight distributions.
Recall that DNN parameters are commonly optimized by the steepest descent optimization algorithm.
To perform the optimization process, we use the Jacobian of the loss function, as in Equation \ref{eq:loss}.\newline

We assume that $W \sim \mathcal{N}(\mu_W,\sigma_W)$, as assumed in \cite{banner2018post}, and to ease the notation, from now on we drop the distribution parameters.

The Jacobian of $\mathcal{L}_p$ does not affect the weight distribution since it is linear with W \cite{lemons2003introduction} 
$$\nabla{\mathcal{L}_p} \propto W \Rightarrow W - \nabla{\mathcal{L}_p} \sim \mathcal{N}$$
Since we want to explain how the Jacobian affects the distribution shift of the weights, we will focus on analyzing the gradients of Equation \ref{eq:kur_loss} w.r.t. W. 
\begin{equation}\label{eq:kur_jacobian}
\frac{\partial \mathcal{L}_K}{\partial W_i} =
\frac{8}{\sigma} \cdot \underbrace{\frac{{(W_i-\mu)}^3}{\sigma^3}}_{\tilde{\mu}_3} \cdot \underbrace{\left|\frac{{(W_i-\mu)}^4}{\sigma^4} - K_T\right|}_{\kappa} 
\end{equation}

\begin{figure}\label{diff_kt}
 	\centering
 	\includegraphics[width=1.\linewidth]{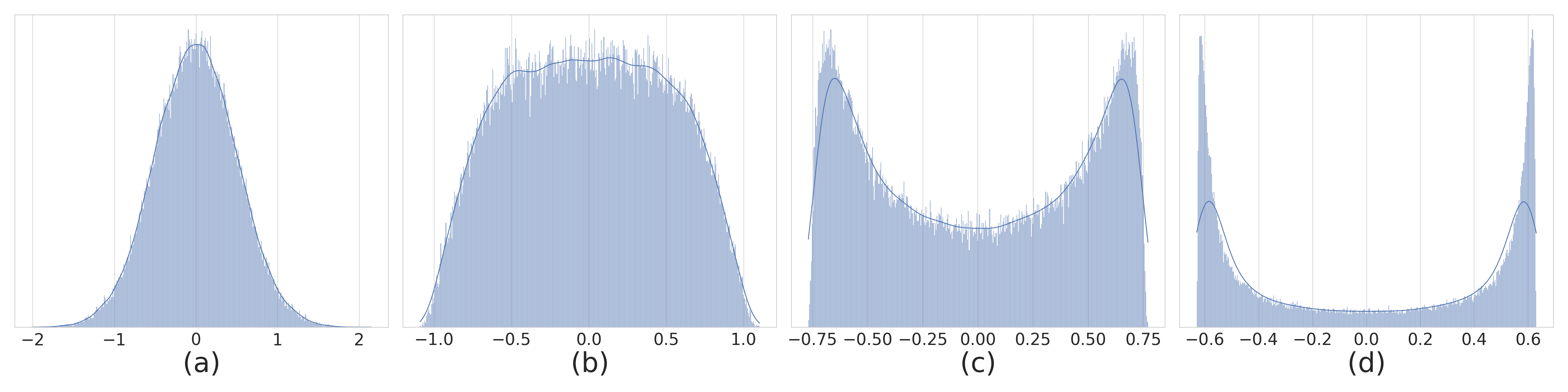}
 	\caption{The distribution density function shift with different values of $K_T$ of one layer with 100,000 weights.From left to right: $K_T=3$ (a), $K_T=1.8$ (b), $K_T=1$ (c) and $K_T=-1.2$ (d). The last one creates a bi-modal distribution that deviates too much and harms the model generalization}
 	\label{fig:Dist_shift_with_kurtosis}
 \end{figure}

Here we derive the partial derivative of the kurtosis loss w.r.t. one layer of the weights $W_i$.
This gradient term consists of a constant coefficient, which does not affect the distribution, and two additional parts:
the third moment, denoted by $\tilde{\mu}_3$, and the distance of the kurtosis from $K_T$, which is denoted by $\kappa$.
$\tilde{\mu}_3$, essentially an odd power of the data, symmetrically shifts weights away from the center of the distribution function. Means, weights with small magnitude, will have gradients with an even smaller magnitude of skewness component. 
$$|W_i| \leq |W_j| \Rightarrow  \tilde{\mu}_3(W_i) \leq (|W_j| - |W_i|) \tilde{\mu}_3(W_j)$$
$\kappa$ holds the desired tail shape of the shifted distribution.\newline
When $K_T=3$, we keep the gradient's distribution normal, and the weight distribution remains normal, as we know for a sum of two random variables with the same $\mu$.
While \cite{chmiel2020robust} holds that for quantization, a uniform distribution is optimal and is achieved by $K_T = 1.8$, we hold that BNNs are more efficient when the weight distribution is bi-modal and symmetrically shifted away from the mean.
The gradient of $\mathcal{L}_K$ is uniform and when $K_T = -1.2$, the weights are shifted uniformly away from the mean.
In practice, this creates a bi-modal distribution after a few epochs, but the generalization error increases and harms the FP32 model. 
An intermediate value, such as $K_T = 1$, might not be a kurtosis value of a well-defined distribution, but can slowly move towards the desired distribution of W.
In Figure \ref{fig:Dist_shift_with_kurtosis}, we show empirically that we end up with a bi-modal distribution, and we do not see any noticeable  harm done to the model generalization.

\begin{figure}[!h]
 	\centering
 	\includegraphics[width=1.05\linewidth]{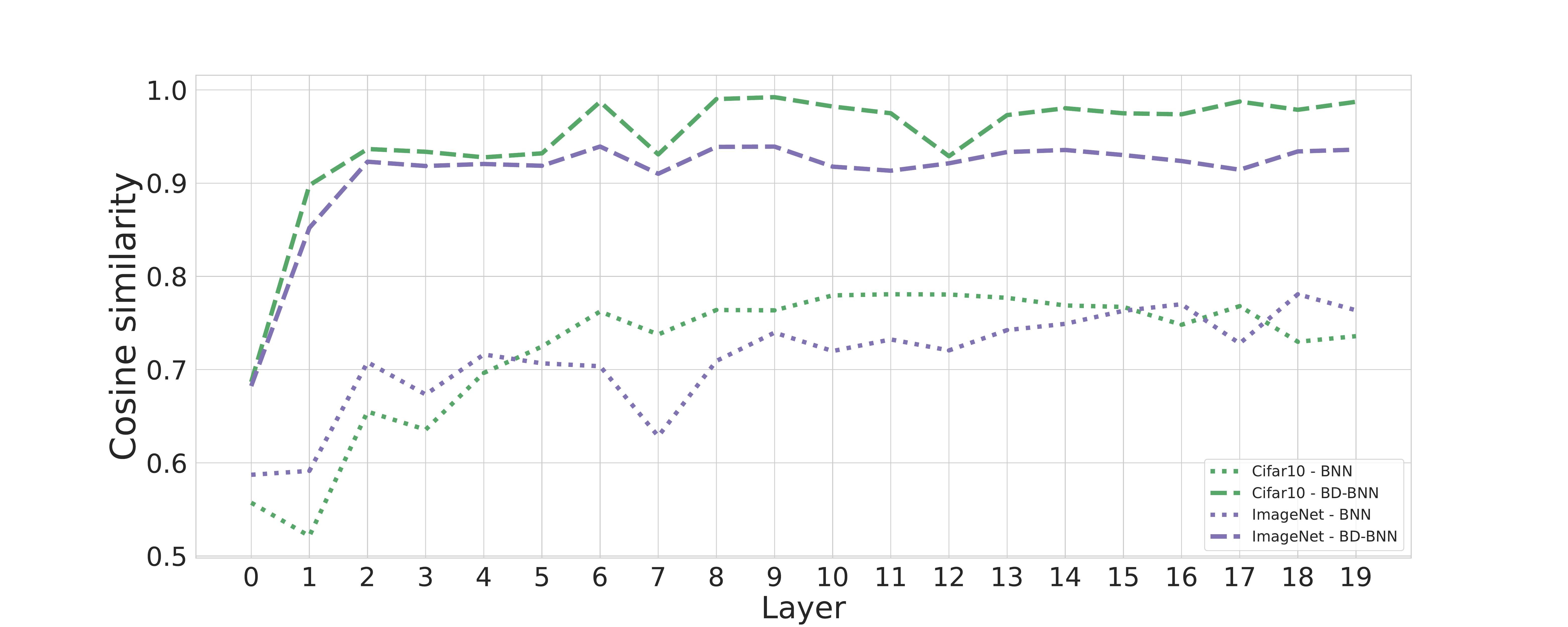}
 	\caption{Cosine similarity per layer of ResNet18, trained on CIFAR10 and ImageNet datasets, dotted lines for simple trained networks, dashed for \methodname{}.}
 	\label{fig:cosine_similarity}
 \end{figure}

In support of this claim, we show in Figure \ref{fig:cosine_similarity} that for $K_T = 1$, the cosine similarity between the FP weights and the binary weights is considerably lower.

As seen in Figure \ref{fig:hetro_kt}, by setting different values of $K_T$ for each layer, we can control the weight distribution of the DNN. We can achieve the desired bi-modal distribution
by setting different $K_T$ for different layers, as long as we ensure that the mean of all $K_T$ values equals the desired one value. We carefully choose them as a set of hyperparameters, so the desired distribution exists in all layers without extending the training process. We use this in order to train a teacher network that will be more sutable for KD with our BNN student. For simplicity, the BNN uses a fixed value $K_T$.

\subsection{Weight Distribution Mimicking}
\label{method:Mimick}
\begin{figure}[!h]
	\centering
	\includegraphics[width=0.9\textwidth]{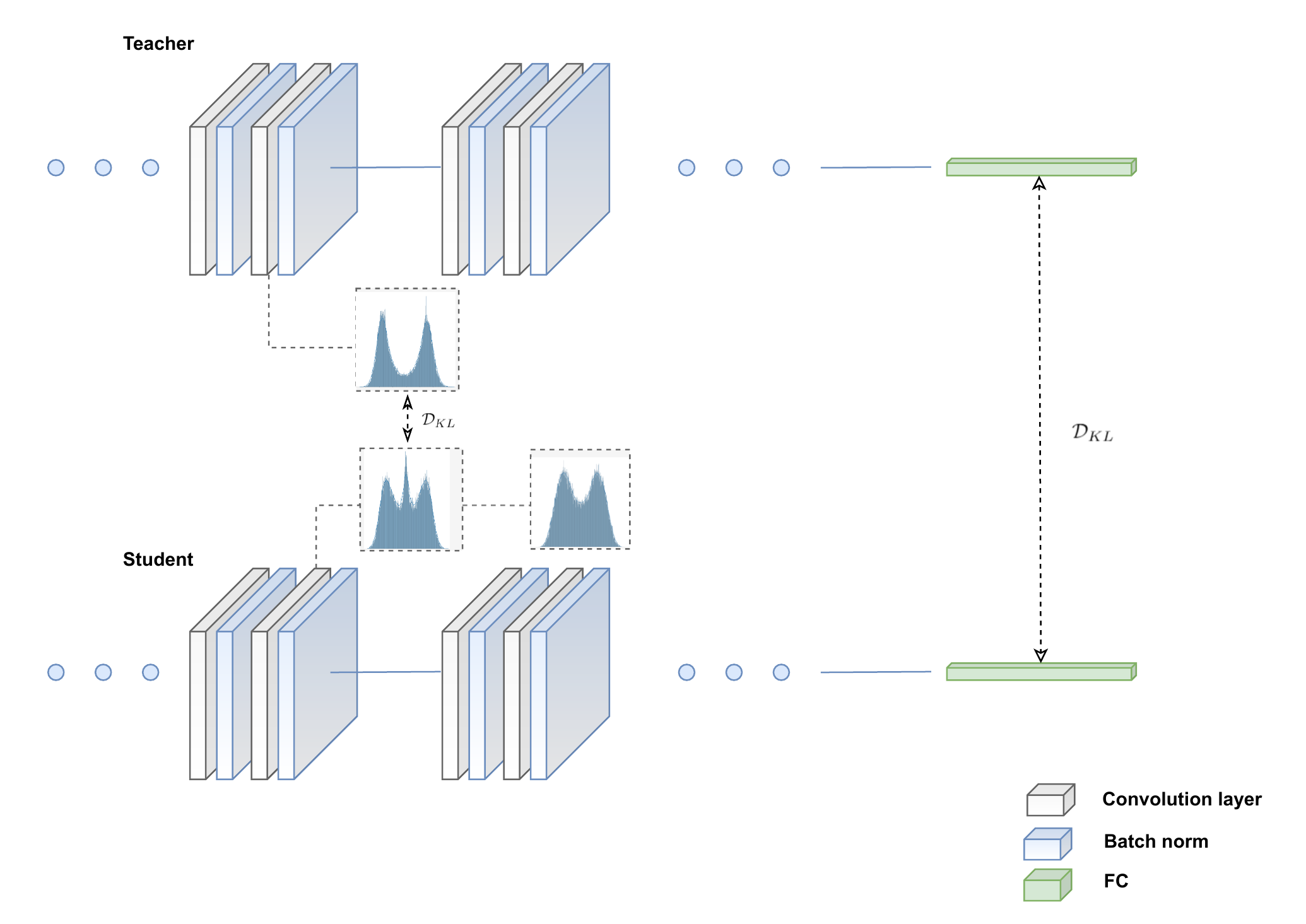} 
	\vspace{-0.3cm}
	\caption{WDM scheme, where we minimize the $\mathcal{L}_{WDM}$ in order to obtain a better bi-modal distribution, as well as KD of the logits between the teacher and the student.}
    \label{fig:wdm}
\end{figure}

The reason BNNs have difficulties obtaining high performance mainly lies in the information loss resulting from the very low precision of weights and activations.
Whereas the quantization loss is tremendous when using ultra-low representation, shifting a FP network and using it as a guide can help minimize this loss.
To bridge this information gap, we propose a teacher-student Weight Distribution Mimicking (WDM) training scheme presented in Figure \ref{fig:wdm}.
WDM uses the information from the FP network teacher to obtain lower loss on the BNN student by minimizing the distance between the distributions.
This method is agnostic to KD and can work in parallel. For that reason we add KD as well and achieve better results.

\begin{equation}\label{eq:WDM_loss}
    \mathcal{L}_{WDM} = \sum_{i=1}^{L} D_{KL} (\mathcal{D}(W_{i,T}) || \mathcal{D}(W_{i,S})) + \beta \mathcal{L}_{KD} (y_T,y_S)
\end{equation}
where we denote the i-th layer of weights in the teacher network as $W_{i,T}$
and the i-th layer of the student BNN as $W_{i,S}$. L is the number of convolution layers, $\mathcal{D}$ is the distribution of the weights, $y_T$ and $y_S$ are the predictions of the teacher and student, respectively, and $D_{KL}$  refers to the Kullback–Leibler divergence. 
The proposed method uses a fixed teacher, meaning we only optimize the student network, and use the teacher as a model towards our desired behavior.
Unlike KD, the goal of WDM is to help guide the BNN to realize how the final distribution of the weights should be at each layer, regardless of the features and predictions.
Moreover, low bit representation networks have difficulties in following the distribution guidance from the kurtosis regularization and require longer training to do so. Using the WDM scheme helps reduce the time it takes to shift the BNN weight distribution. 

\begin{figure}[!h]
	\centering
	\includegraphics[width=0.95\textwidth]{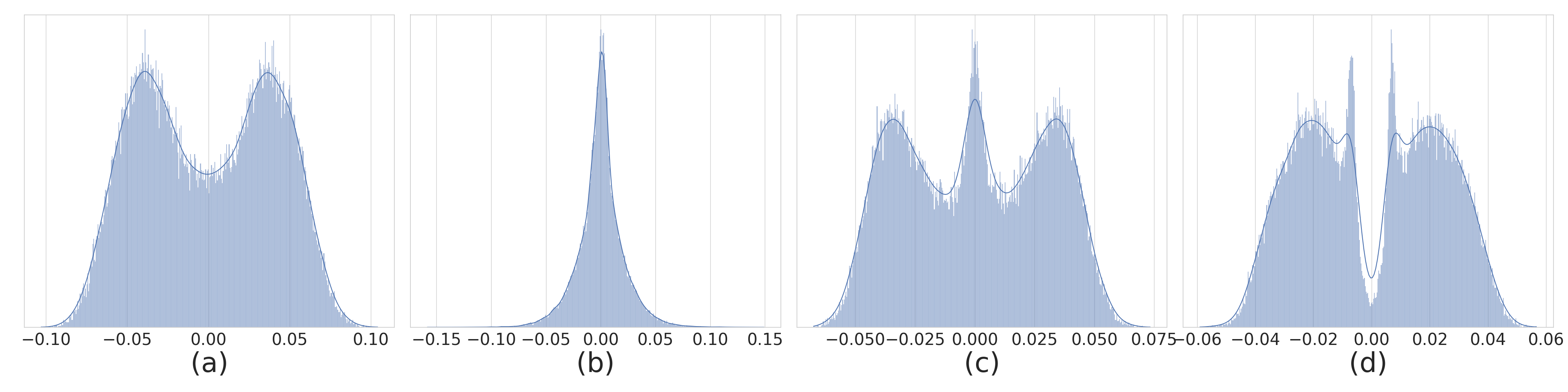} 
	\vspace{-0.3cm}
	\caption{Distribution shifts throughout \methodname{} training. The displayed weights are of the first convolution layer on the second block of ResNet-18 trained on CIFAR-10. (a) is the teacher network, (b) regular BNN, (c) BNN with our weight regularization and (d) WDM applied with teacher-student scheme.}
    \label{fig:bnn_dist}
\end{figure}

\subsection{\methodname{}}
\label{sec:BD-BNN}
Our bi-modal distribution-aware BNN (\methodname{}) is trained in a teacher-student fashion through two main steps: 

(1) We train an FP-32 network and a BNN having the same architecture but with 1-bit representation with weight distribution regularization, where the FP-32 network has different $K_T$ values for each layer, as described in Section \ref{method:regul}. The objective function is Equation \ref{eq:loss} for both networks.

(2) We fix the FP-32 network as a teacher and the BNN as the student in our WDM training scheme, as described in Section \ref{method:Mimick}, and use simple KD \cite{Hinton2015DistillingTK} on top, as shown in Figure \ref{fig:wdm}. We call the student BNN \methodname{}. 
The objective of the \methodname{} in this stage is to aggregate the loss from Equation  \ref{eq:loss} with the one from Equation \ref{eq:WDM_loss} as follows:
\begin{equation}\label{eq:tot_loss}
    \mathcal{L} = \mathcal{L}_{CE} + \lambda\mathcal{L}_K + \alpha\mathcal{L}_{WDM}
\end{equation}

In the \methodname{} optimization process, we use known gradient approximation \cite{Lin2020RotatedBN,Liu2018BiRealNet}.
The weight distribution of \methodname{} shifts at each stage of the training, so the final weight distribution is optimal for the binarization process, as seen in Figure \ref{fig:bnn_dist}. We found that, on average, around 50\% of the weights changed their sign in \methodname{} compared to binarization of a pre-trained network.



%% file: sections/Experiments.tex
\section{Experiments}
In this section we evaluate our \methodname{}{} method on two common image classification datasets, CIFAR-10 \cite{cifar10}  with ResNet-18/20 \cite{resnet} and VGG-small \cite{Simonyan2015VeryDC}, and ImageNet \cite{deng2009imagenet} with ResNet-18 \cite{resnet}.
The first convolution layer and the last fully connected layer are not binarized. We follow the training strategy explained in Section \ref{sec:BD-BNN} and compare our \methodname{} with state-of-the-art existing methods. 

\subsection{CIFAR-10}
For the CIFAR-10 dataset we use an SGD optimizer with a weight decay of 0.001 for better L2 regularization. The learning rate was set to 0.1 and gradually decayed to 0. All convolution layers are binarized, including 1x1 conv. For gradient approximation, we used the method described in \cite{Lin2020RotatedBN}.
We compare our method with several other state-of-the-art binary methods including IR-Net \cite{Qin2019IRNetFA} and RBNN \cite{Lin2020RotatedBN} for the ResNet-18 structure, IR-Net \cite{Qin2019IRNetFA}, RBNN \cite{Lin2020RotatedBN}, FDA-Net \cite{Xu2021LearningFD}, XNOR-Net \cite{Rastegari2016XNORNetIC} and DSQ \cite{Ruihao2018DSQ} for the ResNet-20 structure, and VGG-small. 
In Table \ref{tab:cifar} we compare our \methodname{} method with several existing binarization methods, achieving around 0.2\% accuracy improvement in all the listed architectures. ResNet-20 used the Bi-Real structure that includes a second skip connection in the ResNet block. 

\begin{table}[h!]
    \centering
    \begin{tabular}{@{}c l c c@{}}
    \hline
    Network & Method & Bit-Width (W/A) & Acc(\%) \\
    \hline
    \multirow{4}{7em}{ResNet-18} & FP32 & 32/32 & 93.00 \\
    & IR-Net \cite{Qin2019IRNetFA} & 1/1 & 91.50 \\
    & RBNN \cite{Lin2020RotatedBN}& 1/1 & 92.20 \\
    & \methodname{} (Ours) & 1/1 & \textbf{92.46} \\
    \hline
    \multirow{6}{7em}{ResNet-20} & FP32 & 32/32 & 91.70 \\
    & DoReFa-Net \cite{Shuchang2018DoReFaNet}& 1/1 & 79.30 \\
    & XNOR-Net \cite{Rastegari2016XNORNetIC}& 1/1 & 85.23 \\
    & DSQ \cite{Ruihao2018DSQ}& 1/1 & 84.11 \\
    & IR-Net \cite{Qin2019IRNetFA} & 1/1 & 85.40 \\
    & FDA-BNN \cite{Xu2021LearningFD} & 1/1 & 86.20 \\
    & RBNN \cite{Lin2020RotatedBN}& 1/1 & \textbf{86.50} \\
    & \methodname{} (Ours) & 1/1 & \textbf{86.50}\\
    \hline
    \multirow{6}{7em}{VGG-small} & FP32 & 32/32 & 93.80 \\
    & BNN \cite{Courbariaux2016BinarizedNN} &1/1 & 89.90 \\
    & XNOR-Net \cite{Rastegari2016XNORNetIC}& 1/1 & 89.80 \\
    & DSQ \cite{Ruihao2018DSQ}& 1/1 & 91.72 \\
    & IR-Net \cite{Qin2019IRNetFA}& 1/1 & 90.40 \\
    & RBNN \cite{Lin2020RotatedBN}& 1/1 & 91.30 \\
    & FDA-BNN \cite{Xu2021LearningFD} & 1/1 & 92.54 \\
    & \methodname{} (Ours) & 1/1 & \textbf{92.7} \\
    \hline
    \end{tabular}
    \caption{ Performance comparison with state-of-the-art methods
on CIFAR-10. W/A denotes the bit length of
weights and activations.}
    \label{tab:cifar}
\end{table}

\subsection{ImageNet}
ImageNet is a much more challenging dataset due to its large scale and great
diversity. It contains over 1.2M training images from 1,000 different categories. For ImageNet, we use an ADAM optimizer with a momentum of 0.9 and a learning rate set to $1e^-3$. All methods listed in Table \ref{tab:imagenet}, except \cite{Courbariaux2016BinarizedNN}, do not binarize the down-sampling layer in the shortcut branch of the ResNet block. Similar to \cite{Bulat2019ImpTrain}, we also train the network in two stages. First, we train with binary activations and FP weights, and then we train the full binary network. Furthermore, we use the architecture suggested in \cite{Liu2018BiRealNet} that adds a skip connection to each ResNet block. We use STE gradient approximation as in \cite{Liu2018BiRealNet,Liu2020ReActNetTP}. We compare \methodname{} with IR-Net \cite{Qin2019IRNetFA}, RBNN \cite{Lin2020RotatedBN}, FDA-Net \cite{Xu2021LearningFD}, XNOR-Net \cite{Rastegari2016XNORNetIC}, Bi-real Net \cite{Liu2018BiRealNet}, BNN \cite{Courbariaux2016BinarizedNN} and DoReFa \cite{Shuchang2018DoReFaNet}.  For the ResNet-18 architecture, we show two different settings. The first one is the one described above, and for the second, we use PRelu as the non-linearity function and add a bias to the activations. This setting was used in ReAct-Net \cite{Liu2020ReActNetTP}. In contrast to React-Net \cite{Liu2020ReActNetTP}, however, we keep the ResNet structure as is. We train ReAct-Net with our setting and compare our method to it. 
All experiments are listed in Table \ref{tab:imagenet}. As seen from the table, our \methodname{} achieves 0.5-1.2\% accuracy improvement on ResNet-18.

\begin{table}
    \centering
    \begin{tabular}{@{}c l c c c@{}}
    \hline
    Network & Method & Bit-Width (W/A) & Top-1(\%) & Top-5(\%)\\
    \hline
    \multirow{11}{7em}{ResNet-18} & FP & 32/32 & 69.6 & 89.2\\
    & IR-Net \cite{Qin2019IRNetFA} & 1/1 & 58.1 & 80.0 \\
    & RBNN \cite{Lin2020RotatedBN} & 1/1 & 59.9 & 81.9\\
    & Bi-Real-Net \cite{Liu2018BiRealNet} & 1/1 & 56.4 & 79.5\\
    & XNOR-Net \cite{Rastegari2016XNORNetIC} & 1/1 & 51.2 & 73.2\\
    & BNN \cite{Courbariaux2016BinarizedNN} & 1/1 & 42.2 & 67.1\\
    & DeReFa-Net\cite{Shuchang2018DoReFaNet} & 1/1 & 53.4 & 67.7\\
    & FDA-BNN \cite{Xu2021LearningFD} & 1/1 & 60.2 & 82.3\\
    & \methodname{} (Ours) & 1/1 & \textbf{60.6} & \textbf{82.49}\\
    \cline{2-5}
    & ReActNet* \cite{Liu2020ReActNetTP} & 1/1 & 62.1 & 83.53\\
    & \methodname{}* (Ours) & 1/1 & \textbf{63.27} &  \textbf{84.42}\\
    \hline
    \end{tabular}
    \caption{Performance comparison with state-of-the-art methods on
ImageNet. W/A denotes the bit length of weights
and activations. We report the top-1 and top-5
accuracy performances. '*' indicates that ReAct-Net partial structure was used}
    \label{tab:imagenet}
\end{table}

\subsection{Ablation Study} 
In this section we specify the contribution of each component of the method described in this paper. We test \methodname{} on CIFAR10 with ResNet-20 architecture. As shown in Table \ref{tab:ablation}, each addition contributes to the BNN performance. We conclude from the study that our WDM outperform the commonly used KD training scheme. When put together our method surpasses other state-of-the-art binary methods and achieves highly accurate BNNs.

In addition, we test the effect of the WDM described in Section \ref{method:Mimick}. We try different teacher-student settings on ResNet-18 with the ImageNet dataset. The goal of this ablation is to determine which setting is best for binary networks. We use teachers that were trained with and without weight distribution regularization using the fourth momentum and train a BNN student with a similar structure using the loss specified in Section \ref{method:Mimick}. Table \ref{tab:teacherstudent} shows that the best setting is to train a teacher with kurtosis and let the student try and mimic the teacher's behavior, also using kurtosis. It also shows, as expected, that a teacher and a student that train with different methods show poor results compared to ones that use the same method.

\begin{table}
\caption{Ablation Study of \methodname{}. }
\begin{subtable}[t]{0.48\textwidth}
      \centering
       \begin{tabular}{l c c}
        \toprule
        \textbf{Method} & \textbf{W/A} & \textbf{Acc} \\
        \midrule
        FP & 32/32 & 91.7\% \\
        BNN & 1/1 & 83.9\% \\
        BNN + WDR & 1/1 & 85.25\% \\
        BNN + WDR + KD & 1/1 & 85.69\% \\
        BNN + WDR + WDM & 1/1 & 86.50\% \\
        \bottomrule
       \end{tabular}
      \caption{Exploration of \methodname{}. WDR and WDM stands for weight distribution regularization and weight distribution mimicking respectively.}
      \label{tab:ablation}
    \end{subtable}%
    \hfill
\begin{subtable}[t]{0.48\textwidth}
      \centering
        \begin{tabular}{ p{2cm} p{2cm} p{1cm} }
        \toprule
        \textbf{Teacher} & \textbf{Student} & \textbf{Acc}\\
        \midrule
        FP & BNN & 58.45\% \\
        FP + WDR & BNN & 56.32\% \\
        FP & BNN + WDR & 55.25\% \\
        FP + WDR & BNN + WDR & 60.5\% \\
        \bottomrule
        \end{tabular}
        \caption{WDM scheme analysis with different training of teacher-student.}
      \label{tab:teacherstudent}
    \end{subtable}%
\end{table}


%% file: sections/Conclusions.tex
\section{Conclusions}
Binarization is one of the dominant methods to reduce the inference complexity of DNNs, with the ability to reduce the memory footprint by 32x and to accelerate the deployment by 58x \cite{Rastegari2016XNORNetIC}. Although this promising compression ability, there is still a gap between the accuracy of the binarize network and their full precision counterpart.

In this work, we go one step forward in reducing this gap and presented a novel method for binarization-aware training. Our approach is based on manipulating the weight distribution of the binary neural network for better performance. We show that a bi-model distribution achieves better results and higher cosine similarity than common binarization methods. We incorporate the fourth-moment weight regularization to the loss function. As a result, the weight distribution shifts and is more suitable for binary models. In addition, we presented a new teacher-student weight mimicking method that helps minimize the binarization error. Those two components form our presented \methodname{} that outperforms state-of-the-art BNNs by up to 1.2\% top-1 accuracy on ImageNet.
